\documentclass{article}

     \PassOptionsToPackage{numbers, compress}{natbib}

\usepackage[final]{neurips_2024}

\usepackage{eccvabbrv}
\usepackage{graphicx}
\usepackage{enumitem}
\usepackage{xfrac}
\usepackage{makecell}
\usepackage{multirow}
\usepackage{amsmath}
\usepackage[normalem]{ulem}
\useunder{\uline}{\ul}{}

\usepackage[utf8]{inputenc} 
\usepackage[T1]{fontenc}    
\usepackage{hyperref}       
\usepackage{url}            
\usepackage{booktabs}       
\usepackage{amsfonts}       
\usepackage{nicefrac}       
\usepackage{microtype}      
\usepackage{xcolor}         

\title{DCDepth: Progressive Monocular Depth Estimation\\in Discrete Cosine Domain}

\author{
    Kun Wang\textsuperscript{\rm 1}, Zhiqiang Yan\textsuperscript{\rm 1}, Junkai Fan\textsuperscript{\rm 1}, Wanlu Zhu\textsuperscript{\rm 1}, Xiang Li\textsuperscript{\rm 2}, Jun Li\textsuperscript{\rm 1}\thanks{Corresponding authors.}\ \ and Jian Yang\textsuperscript{\rm 1}\footnotemark[1]
    \\
    \textsuperscript{\rm 1}PCA Lab, Nanjing University of Science and Technology, China\\
    \textsuperscript{\rm 2}Nankai University, China\\
}

\begin{document}

\maketitle

\begin{abstract}
    In this paper, we introduce DCDepth, a novel framework for the long-standing monocular depth estimation task. Moving beyond conventional pixel-wise depth estimation in the spatial domain, our approach estimates the frequency coefficients of depth patches after transforming them into the discrete cosine domain. This unique formulation allows for the modeling of local depth correlations within each patch. Crucially, the frequency transformation segregates the depth information into various frequency components, with low-frequency components encapsulating the core scene structure and high-frequency components detailing the finer aspects. This decomposition forms the basis of our progressive strategy, which begins with the prediction of low-frequency components to establish a global scene context, followed by successive refinement of local details through the prediction of higher-frequency components. We conduct comprehensive experiments on NYU-Depth-V2, TOFDC, and KITTI datasets, and demonstrate the state-of-the-art performance of DCDepth. Code is available at \url{https://github.com/w2kun/DCDepth}.
\end{abstract}

\section{Introduction}

\textbf{M}onocular \textbf{D}epth \textbf{E}stimation (MDE) is a cornerstone topic within computer vision communities, tasked with predicting the distance--or depth--of each pixel’s corresponding object from the camera based solely on single image. As a pivotal technology for interpreting 3D scenes from 2D representations, MDE is extensively applied across various fields such as autonomous driving, robotics, and 3D modeling \cite{depth_driving, rignet, depth_robotics, altnerf}, \etc. However, MDE is challenged by the inherent ill-posed nature of inferring 3D structures from 2D images, making it a particularly daunting task for traditional methodologies, which often hinge on particular physical assumptions or parametric models \cite{depth_from_perspective, shape_from_shading, depth_mrf, saxena2007depth}.

Over the past decade, the field of computer vision has witnessed a substantial surge in the integration of deep learning techniques. Many studies have endeavored to harness the robust learning capabilities of end-to-end deep neural networks for MDE task, propelling the estimation accuracy to new heights. Researchers have investigated a variety of methodologies, including regression-based \cite{eigen_depth, megadepth, vnl}, classification-based \cite{cao2017estimating, fu2018deep}, and classification-regression based approaches \cite{adabins, binsformer}, to predict depth on a per-pixel basis within the spatial domain. Despite these significant strides in enhancing accuracy, current methods encounter two primary limitations: the first is the tendency to predict depth for individual pixels in isolation, thus neglecting the crucial local inter-pixel correlations. The second limitation is the reliance on a singular forward estimation process, which may not sufficiently capture the complexities of 3D scene structures, thereby constraining their predictive performance.

\begin{figure*}
    \centering
    \includegraphics[width=0.9\linewidth]{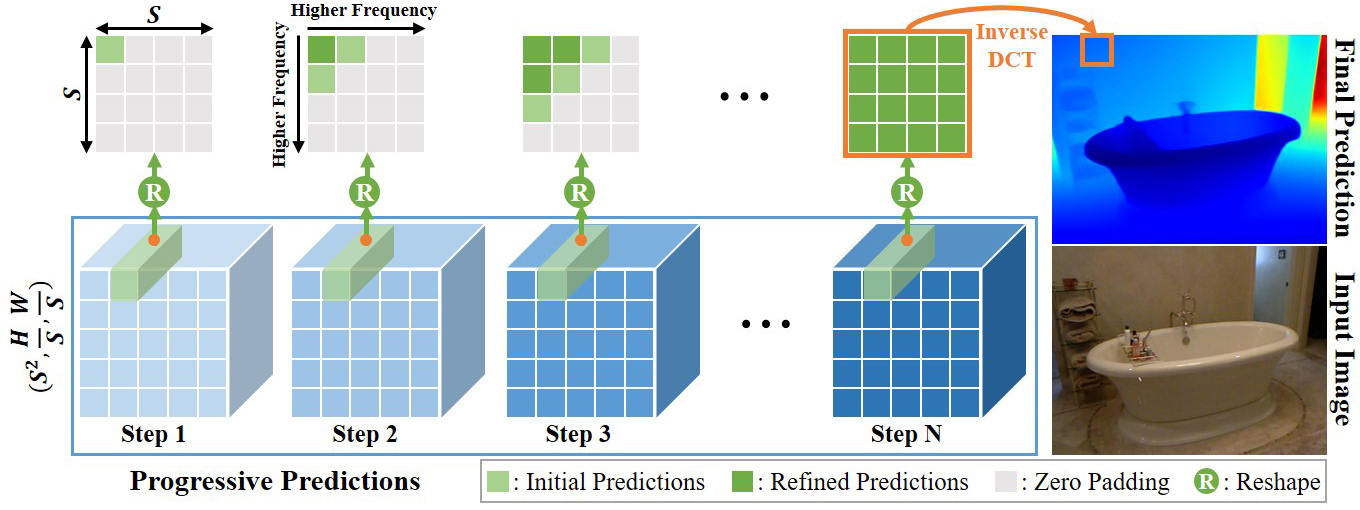}
    \vskip -0.05in
    \caption{\textbf{Progressive estimation scheme.} For input image with size $H\times W$, DCDepth estimates the DCT coefficients for each $S\times S$ depth patches. The prediction follows a global-to-local strategy, starting with the initial estimation of lower-frequency components to capture the global scene structure. Subsequently, higher-frequency components  are estimated to enhance the local details, while the lower-frequency estimates are refined. The estimation is carried out at $\frac{H}{S}\times \frac{W}{S}$ resolution, and spatial-domain estimation is achieved through inverse DCT.}
    \label{fig.1}
    \vskip -0.05in
\end{figure*} 

To address the identified limitations, we propose to transfer depth estimation from the spatial domain to the frequency domain. Instead of directly predicting metric depth values, our method focuses on estimating the frequency coefficients of depth patches transformed using the \textbf{D}iscrete \textbf{C}osine \textbf{T}ransform (DCT) \cite{dct, fast_dct}. This strategy offers dual benefits: firstly, the DCT’s basis functions inherently capture the inter-pixel correlations within depth patches, thereby facilitating the model’s learning of local structures. Secondly, the DCT decomposes depth information into distinct frequency components, where low-frequency components reflect the overall scene architecture, and high-frequency components capture intricate local details. This dichotomy underpins our progressive estimation methodology, which commences with the prediction of low-frequency coefficients to grasp the macroscopic scene layout, subsequently refining the local geometries by inferring higher-frequency coefficients predicated on previous predictions. The spatial depth map is then accurately reconstructed via the inverse DCT. We illustrate this progress in Fig. \ref{fig.1}. To implement our progressive estimation, we introduce a \emph{Progressive Prediction Head} (PPH) that conditions on previous predictions from both spatial and frequency domains, and facilitates the sequential prediction of higher-frequency components using a GRU-based mechanism. Furthermore, recognizing the DCT’s energy compaction property--indicative of the concentration of signal data within low-frequency components--we introduce a DCT-inspired downsampling technique to mitigate information loss during the downsampling process. This technique is embedded within a \emph{Pyramid Feature Fusion} (PFF) module, ensuring effective fusion of multi-scale image features for accurate depth estimation.

Our contributions can be succinctly summarized in three key aspects:

\begin{itemize}[itemsep=0.0em, topsep=-0.1em, leftmargin=2.0em]
    \item To the best of our knowledge, we are the first to formulate MDE as a progressive regression task in the discrete cosine domain. Our proposed method not only models local correlations effectively but also enables global-to-local depth estimation.
    \item We introduce a framework called DCDepth, comprising two novel modules: the PPH module progressively estimates higher-frequency coefficients based on previous predictions, and the PFF module incorporates a DCT-based downsampling technique to mitigate information loss during downsampling and ensures effective integration of multi-scale features.
    \item We evaluate our approach through comprehensive experiments on NYU-Depth-V2 \cite{nyu}, TOFDC \cite{tofdc}, and KITTI \cite{kitti} datasets. The results demonstrate the superior performance of DCDepth compared to existing state-of-the-art methods.
\end{itemize}

\section{Related Work}

\textbf{M}onocular \textbf{D}epth \textbf{E}stimation (MDE) remains a central theme in computer vision, essential for translating 2D imagery into 3D scene geometry. The evolution of MDE has been markedly influenced by the integration of deep neural networks. A foundational advancement was introduced by Eigen et al. \cite{eigen_depth}, who developed a multi-scale deep convolutional network architecture, comprising a global network for coarse depth prediction and a local network for refinement. They also introduced a scale-invariant loss function to address the scale ambiguity challenge inherent in MDE. Building on this, subsequent researches \cite{megadepth, yin2021learning, rnw, desnet, yan2023distortion, yu2023aggregating, yan2024learnable} have adopted end-to-end regression approaches with deep convolutional networks to further tackle MDE’s challenges.

However, inferring depth from a single image is intrinsically problematic due to the countless potential depth maps that can correspond to one image. To mitigate this, additional information and constraints have been incorporated into the MDE task, such as semantics \cite{semantic_depth, panet} and surface normals \cite{geonet, nddepth}. Further enhancements in depth estimation accuracy have been achieved through attention mechanisms \cite{hao2018detail, xu2018structured, dpt}, multivariate gaussian modeling \cite{mgdepth}, internal discretization technique \cite{idisc} and pretraining \cite{yan2022multi, xie2023revealing}.
In contrast to the regression-based approach, some works \cite{fu2018deep, cao2017estimating} have conceptualized MDE as a classification task, estimating the probability distribution of depth values. Yet, these methods often produce discontinuities due to discrete depth outputs. To overcome this, alternative strategies \cite{adabins, binsformer, localbins, iebins} have combined classification and regression formulations, learning probabilistic distributions and employing linear combinations with depth candidates for final depth predictions. Our methodology diverges from these paradigms by progressively estimating frequency coefficients for depth patches after their transformation into the discrete cosine domain. This approach not only enhances computational efficiency but also achieves state-of-the-art performance.

\section{Method}

In this section, we introduce our progressive depth estimation framework, DCDepth. We begin by providing an overview of the 2D \textbf{D}iscrete \textbf{C}osine \textbf{T}ransform (DCT) as essential background knowledge. Subsequently, we delve into the progressive estimation scheme and elaborate on the network architecture. Finally, we present the loss function employed for training our model.

\subsection{Reviewing 2D Discrete Cosine Transform}
\label{sec.dct}

The 2D DCT is a mathematical technique used to decompose 2D discrete signals, such as depth maps and feature maps, into a sum of cosine basis functions with varying frequencies. The basis functions are defined as follows:
\begin{equation}
    B^{i,j}_{u,v} = \alpha(u) \alpha(v) \cos \left[ \frac{\pi}{W} \left(i + \frac{1}{2} \right) u \right] \cos \left[ \frac{\pi}{H} \left(j + \frac{1}{2} \right) v \right],
\end{equation}
where $u\in \left[0, W-1\right]$ and $v\in \left[0, H-1\right]$ represent the frequency indices, $i\in \left[0, W-1\right]$ and $j\in \left[0, H-1\right]$ denote the signal indices, and $W$ and $H$ indicate the input resolution. The terms $\alpha(u)$ and $\alpha(v)$ correspond to normalization factors. The forward process of 2D DCT, denoted as $T(\cdot)$, transforms the input signal $x\in\mathbb{R}^{H\times W}$ in the spatial domain to the frequency spectrum $f=T(x), f\in\mathbb{R}^{H\times W}$, and can be expressed as:
\begin{equation}
    f_{u,v}=\sum^{W-1}_{i=0}\sum^{H-1}_{j=0}x_{i,j}B^{i,j}_{u,v}.
    \label{eq.dct}
\end{equation}
The resulting $f$ is a matrix with the same size as the input $x$, with low-frequency components located near the top-left corner and high-frequency components near the bottom-right corner. The upper left one with zero frequency is called the DC components, and the remains are AC components. Low-frequency components typically characterize smooth regions, while high-frequency components capture edges or fine details where signal values change rapidly. The inverse 2D DCT, denoted as $T^{-1}(\cdot)$, performs the reverse operation by transforming the frequency spectrum $f$ back to the spatial domain $x=T^{-1}(f)$, and can be formulated as:
\begin{equation}
    x_{i,j}=\sum^{W-1}_{u=0}\sum^{H-1}_{v=0}f_{u,v}B^{i,j}_{u,v}.
    \label{eq.inv_dct}
\end{equation}
The DCT has two desirable advantages. Firstly, it operates in the real number domain, simplifying the data processing. Secondly, it exhibits superior energy compaction properties by concentrating the majority of information within a small number of low-frequency components.

\subsection{Progressive Estimation in Discrete Cosine Domain}
\begin{figure*}
    \centering
    \includegraphics[width=0.95\linewidth]{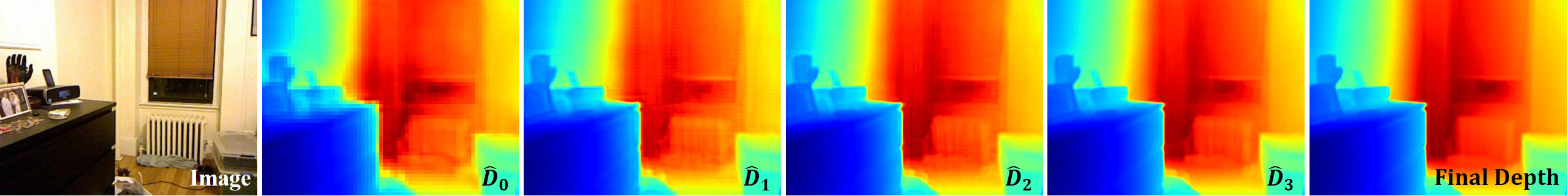}
    \vskip -0.05in
    \caption{\textbf{Evolution of intermediate depth estimations.} We report several intermediate depth estimation results to illustrate our progressive estimation scheme.}
    \label{fig.progressive}
    \vskip -0.05in
\end{figure*}

Estimating depth from a single image remains a challenging task, particularly for scenes with intricate geometry. To tackle this, we propose a progressive method based on 2D DCT to estimate scene depth progressively from a global perspective down to local details. The entire process is illustrated in Fig. \ref{fig.1}. We denote the input image as $\mathcal{I}\in\mathbb{R}^{3\times H\times W}$. Our proposed method, symbolized as $\Psi(\cdot)$, predicts the frequency coefficients $\mathcal{C}\in\mathbb{R}^{S^2\times\frac{H}{S}\times\frac{W}{S}}$ for non-overlapping depth patches $\mathcal{P}\in\mathbb{R}^{S\times S}$, where $S$ is set to 8 in our framework. These coefficients are subsequently transformed back to the spatial domain $\hat{\mathcal{D}}\in\mathbb{R}^{H\times W}$ using the inverse 2D DCT, as expressed by
\begin{equation}
    \hat{\mathcal{D}} = T^{-1}(\Psi(\mathcal{I})).
\end{equation}
The separation of low- and high-frequency components in a depth map effectively divides the scene into overall structures with gradual depth changes and local details with sharp depth transitions. This frequency characteristic enables us to break down the challenging MDE task into multiple prediction stages, progressing from simpler to more complex predictions. Initially, the DC coefficient $\mathcal{C}_0$ is predicted, establishing a foundational depth context. Subsequently, the AC coefficients $\{\mathcal{C}_i\}_{i=1}^{S^2-1}$ are iteratively estimated in ascending frequency order. During the inverse transformation to the spatial domain, any coefficients yet to be predicted are padded with zeros. In each iterative step $k$, we not only predict higher-frequency components but also refine the preceding frequency predictions
\begin{equation}
    \mathcal{C}^{k}=\mathcal{C}^{k-1}+\Delta\mathcal{C}^{k},
\end{equation}
by estimating a correction term $\Delta\mathcal{C}^{k}$. To reduce the required iterations for estimating all $S^2$ coefficients, we utilize the energy compaction property of DCT, and partition the frequency spectrum $\mathcal{C}$ into subgroups along the subdiagonal, yielding $2S-1$ subgroups $\{g_i\}^{2S-1}_{i=0}$. By merging the high-frequency subgroups, we further streamline the iterative process. This grouping strategy ensures that lower-frequency groups contain fewer components necessitating more prediction steps, while higher-frequency groups encompass a larger number of components requiring fewer steps. The intermediate depth maps are provided in Fig. \ref{fig.progressive} to elucidate the step-by-step prediction process.

\subsection{DCDepth Architecture}

\begin{figure*}
    \centering
    \includegraphics[width=0.9\linewidth]{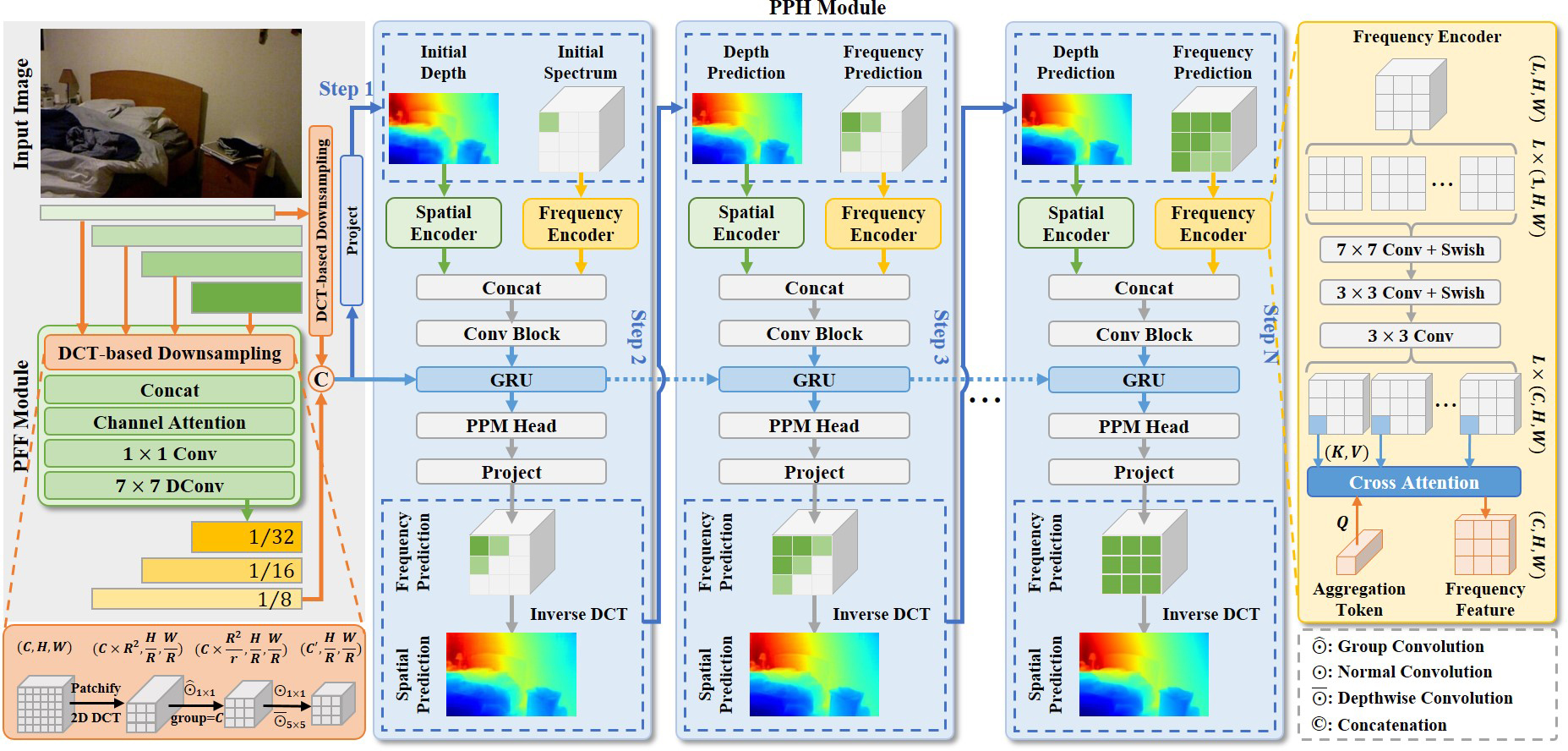}
    \vskip -0.05in
    \caption{\textbf{DCDepth framework overview.} The DCT-based downsampling strategy is shown at the bottom-left corner, where $R$ and $r$ denote for downsampling factor and channel reduction rate, respectively. The central section details the iterative process of PPH, with $N$ indicating the number of iterative steps. The frequency encoder utilized by PPH is illustrated at the right box.}
    \label{fig.framework}
    \vskip -0.02in
\end{figure*}

\paragraph{Overview} We present the comprehensive framework of DCDepth in Fig. \ref{fig.framework}, which comprise four key components: an image encoder, a \textbf{P}yramid \textbf{F}eature \textbf{F}usion (PFF) module, a decoder, and a \textbf{P}rogressive \textbf{P}rediction \textbf{H}ead (PPH). The image encoder acts as a robust feature extractor capturing image features $\mathcal{F}=\{\mathcal{F}_0, \mathcal{F}_1, \mathcal{F}_2, \mathcal{F}_3\}$ at varying resolutions of $\sfrac{1}{4}$, $\sfrac{1}{8}$, $\sfrac{1}{16}$, and $\sfrac{1}{32}$ relative to the input image size. These multi-scale features are advantageous as the shallow features contain texture-related details, while the deep features hold global and semantic information essential for scene understanding. The PFF module, symbolized as $\Gamma(\cdot)$, is introduced to effectively amalgamate these features, yielding a comprehensive integrated feature representation $\mathcal{F}'=\Gamma(\mathcal{F})$. The decoder, denoted as $D(\cdot)$, consists of three neural CRF \cite{newcrf} modules and two PixelShuffle \cite{pixelshuffle} modules. This configuration processes and upscales $\mathcal{F}'$ to $\hat{\mathcal{F}}=D(\mathcal{F}')$, achieving $\sfrac{1}{8}$ of the original resolution. The PPH performs estimations at the same resolution as $\hat{\mathcal{F}}$. It begins by down-sampling $\mathcal{F}_0$ to half its resolution using the proposed DCT-based downsampling. This down-sampled feature is then concatenated with $\hat{\mathcal{F}}$, forming the initial hidden state for the progressive estimation.

\paragraph{Pyramid Feature Fusion Module} The primary objective of PFF is to harness the wealth of information embedded in the multi-scale image features, thereby creating a more comprehensive and enriched feature representation conducive to scene understanding. The layout of PFF is depicted in the left box of Fig. \ref{fig.framework}. Effective feature aggregation necessitates a proficient downsampling strategy to mitigate information loss, especially when downscaling at larger magnifications. To address this, we introduce a novel DCT-based downsampling strategy engineered to minimize information loss during downsampling.
The operational procedure of this strategy is elucidated in the bottom-left corner of Fig. \ref{fig.framework}. Consider a feature map $\mathcal{F}\in\mathbb{R}^{C\times H\times W}$ slated for downsampling by a factor of $R$. We begin by partitioning $\mathcal{F}$ into patches $\mathcal{P}\in\mathbb{R}^{C\times R^2\times\frac{H}{R}\times\frac{W}{R}}$. Each channel of $\mathcal{P}$ is then individually subjected to Eq. \ref{eq.dct} to transform the feature maps into the frequency domain. Leveraging the energy compaction property of the DCT, the key information within $\mathcal{F}$ is condensed into a few dominant frequency components characterized by large absolute values. This compression enables us to selectively reduce the number of channels from $C\times R^2$ to $C\times\frac{R^2}{r}$ with a reduction rate of $r$ via $1\times1$ convolutions configured with groups set to $C$. The squeezed feature maps are then consolidated through a sequence of operations involving a $1\times1$ convolution followed by a $5\times5$ depth-wise convolution, culminating in the generation of the final output featuring $C'$ channels and reduced spatial resolution.

\paragraph{Progressive Prediction Head} The PPH, as depicted in the middle segment of Fig. \ref{fig.framework}, incorporates two specialized encoders: $E_s(\cdot)$ for spatial-domain inputs and $E_f(\cdot)$ for frequency-domain inputs. The spatial encoder, composed of three convolutional layers with a stride of 2, convolves and downsamples the spatial-domain input $\hat{\mathcal{D}}$, producing a feature map at $\sfrac{1}{8}$ of the original resolution. The architecture of $E_f(\cdot)$ is outlined in the right box of Fig. \ref{fig.framework}. For frequency input $\mathcal{C}\in\mathbb{R}^{L\times H\times W}$, where $L$ signifies the number of valid frequency components, we first split them into $L$ chunks with shape $1\times H\times W$. Each chunk is then processed through three convolutional layers with Swish activation \cite{swish} to extract features of dimensions $C\times H\times W$ for each frequency component. Given the variability in the number of valid frequency components across different iterative steps, we employ cross-attention \cite{transformer, vit} mechanism to merge information from the various frequency components. A learnable \emph{aggregation token} of dimensions $1\times C$ is introduced to compile information from individual frequency components at each pixel location, yielding feature outputs of shape $C\times H\times W$ and effectively compressing the dimension $L$. The PPH operates iteratively, utilizing a \textbf{G}ated \textbf{R}ecurrent \textbf{U}nit (GRU) \cite{gru, raft}, denoted as $G(\cdot,\cdot)$, to encode the historical estimation states
\begin{equation}
    \mathcal{H}_i=G(E_s(\hat{\mathcal{D}}_{i-1}),E_f(\mathcal{C}_{i-1})),
\end{equation}
prior to iterative step $i$. The hidden state $\mathcal{H}$ is then projected to the coefficient output by a Pyramid Pooling Module (PPM) \cite{ppm} to aggregate global context, followed by a linear projection.

\subsection{Loss Function}

We employ the scaled scale-invariant loss \cite{bts, adabins} to calibrate the model’s depth estimations $\hat{\mathcal{D}_i}$ at each iterative step $i$ against the ground truth depth map $\mathcal{D}$. The loss function is formulated as:
\begin{equation}
    L_d = \alpha\cdot\sum_{i=1}^{N} \beta^{N-i} \sqrt{\frac{1}{M} \sum d_i^2 - \frac{\lambda}{M^2}(\sum d_i)^2},
\end{equation}
where $d = \hat{\mathcal{D}_i} - \mathcal{D}$, $N$ denotes the number of iterative steps, and $M$ represents the number of valid depth values. We consistently set $\alpha=10$, $\beta=0.8$ and $\lambda = 0.85$ across all experiments. The presence of missing values in the depth ground truth can render the model’s frequency-domain predictions inadequately supervised. To mitigate this, we introduce two regularization terms. Specifically, to enforce the sparsity of high-frequency coefficients, we define the frequency regularization loss as:
\begin{equation}
    L_f = \sum (\epsilon^{u+v} - 1)\cdot \vert f_{u,v}\vert,
\end{equation}
where $f_{u,v}$ is the frequency coefficient indexed by $(u,v)$, and $\epsilon$ is set to $1.2$. Additionally, we incorporate a smoothness term to promote the smoothness of $\hat{\mathcal{D}}$:
\begin{equation}
    L_s=\vert\partial_x \hat{\mathcal{D}}\vert\cdot e^{-\vert\partial_x I_t\vert}+\vert\partial_y \hat{\mathcal{D}}\vert\cdot e^{-\vert\partial_y I_t\vert},
\end{equation}
where $\partial_x$ and $\partial_y$ represent image gradient along horizontal and vertical axes, respectively, and $\vert\cdot\vert$ denote the absolute value function. The final loss is the weighted summation of these three loss terms.

\section{Experiment}

\begin{table*}
    \centering
    \footnotesize
    \renewcommand\arraystretch{1.25}
    \resizebox{0.95\linewidth}{!}{
        \begin{tabular}{c||c||cccc||ccc}
            \Xhline{1.5pt}
            Method        & Backbone      & Abs Rel $\downarrow$ & Sq Rel $\downarrow$ & RMSE $\downarrow$ & $\text{log}_{10}$ $\downarrow$ & $\delta<1.25$ $\uparrow$ & $\delta<1.25^2$ $\uparrow$ & $\delta<1.25^3$ $\uparrow$ \\ \hline\hline
            DORN \cite{fu2018deep}         & ResNet-101    & 0.115          & –              & 0.509          & 0.051             & 0.828          & 0.965           & 0.992           \\
            VNL \cite{vnl}                 & ResNet-101    & 0.108          & –              & 0.416          & 0.048             & 0.875          & 0.976           & 0.994           \\
            BTS \cite{bts}                 & DenseNet-161  & 0.110          & 0.066          & 0.392          & 0.047             & 0.885          & 0.978           & 0.994           \\
            ASNDepth \cite{asndepth}       & HRNet-48      & 0.101          & –              & 0.377          & 0.044             & 0.890          & 0.982           & 0.996           \\
            TransDepth \cite{transdepth}   & R-50+ViT-B/16 & 0.106          & –              & 0.365          & 0.045             & 0.900          & 0.983           & 0.996           \\
            AdaBins \cite{adabins}         & E-B5+mini-ViT & 0.103          & –              & 0.364          & 0.044             & 0.903          & 0.984           & {\ul 0.997}     \\
            LocalBins \cite{localbins}     & E-B5          & 0.099          & –              & 0.357          & 0.042             & 0.907          & 0.987           & \textbf{0.998}  \\ \hline
            NeWCRFS \cite{newcrf}          & Swin-Large    & 0.095          & 0.045          & 0.334          & 0.041             & 0.922          & \textbf{0.992}  & \textbf{0.998}  \\
            BinsFormer \cite{binsformer}   & Swin-Large    & 0.094          & –              & 0.330          & 0.040             & 0.925          & 0.989           & {\ul 0.997}     \\
            PixelFormer \cite{pixelformer} & Swin-Large    & 0.090          & –              & 0.322          & 0.039             & 0.929          & {\ul 0.991}     & \textbf{0.998}  \\
            IEBins \cite{iebins}           & Swin-Large    & 0.087          & {\ul 0.040}    & 0.314          & {\ul 0.038}       & 0.936          & \textbf{0.992}  & \textbf{0.998}  \\
            MG-Depth \cite{mgdepth}          & Swin-Large    & 0.087          & –              & {\ul 0.311}    & –                 & 0.933          & –               & –               \\
            NDDepth \cite{nddepth}               & Swin-Large    & 0.087          & 0041           & {\ul 0.311}    & {\ul 0.038}       & 0.936          & {\ul 0.991}     & \textbf{0.998}  \\
            VA-DepthNet \cite{vadepth}     & Swin-Large    & {\ul 0.086}    & \textbf{0.039} & \textbf{0.304} & \textbf{0.037}    & {\ul 0.937}    & \textbf{0.992}  & \textbf{0.998}  \\ \hline
            \textbf{Ours}                  & Swin-Large    & \textbf{0.085} & \textbf{0.039} & \textbf{0.304} & \textbf{0.037}    & \textbf{0.940} & \textbf{0.992}  & \textbf{0.998}  \\
            \Xhline{1.5pt}
        \end{tabular}
    }
    \vskip -0.05in
    \caption{\textbf{Quantitative depth comparison on NYU-Depth-V2 dataset.} The maximum depth is capped at 10 meters. R-50 and E-B5 represent for ResNet-50 \cite{resnet} and EfficientNet-B5 \cite{efficientnet}, respectively. '-' means not applicable. The best result is in \textbf{bold}, and the second is \uline{underlined}.}
    \label{tab.nyu}
    \vskip -0.05in
\end{table*}

\begin{figure*}
    \centering
    \includegraphics[width=0.95\linewidth]{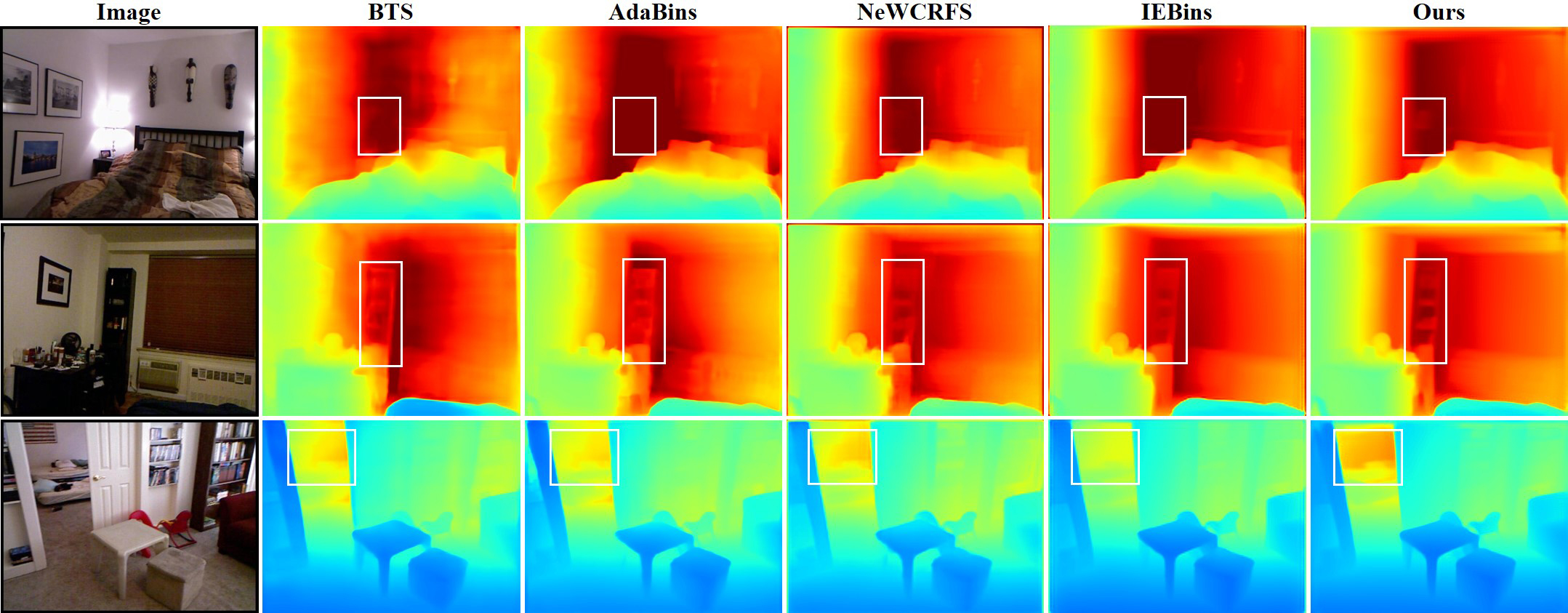}
    \vskip -0.05in
    \caption{\textbf{Qualitative depth comparison on the NYU-Depth-V2 dataset.} The white boxes highlight the regions where our method achieves more accurate predictions.}
    \label{fig.nyu}
    \vskip -0.05in
\end{figure*}

In this section, we evaluate DCDepth by conducting a comparative analysis with established methodologies. We commence by delineating the datasets and evaluation metrics employed in our evaluation. Subsequently, we detail the implementation specifics that underpin our experiments. Concluding this section, we demonstrate the efficacy of the proposed modules via extensive ablation studies.

\subsection{Dataset and Evaluation Metric}

\paragraph{Dataset} We evaluate our method on three datasets that covers a diverse array of indoor and outdoor scenes.
\textbf{(1) NYU-Depth-V2} \cite{nyu} is centered on indoor environments and consists of RGB-D images captured with a Microsoft Kinect sensor. The settings span various indoor scenes such as bedrooms, offices, and classrooms. The images in this dataset are presented at a resolution of $640\times 480$. We follow the data split as outlined in BTS \cite{bts}, featuring 24231 training images and 654 test images.
\textbf{(2) TOFDC} \cite{tofdc} is collected using a mobile phone paired with a lightweight Time-of-Flight (ToF) camera, capturing a wide array of subjects like flowers, human figures, and toys under different scenes and lighting conditions. The dataset is divided into 10,000 training samples and 560 testing samples, with images at a resolution of $512\times 384$. 
\textbf{(3) KITTI} \cite{kitti} is a well-known outdoor dataset that features RGB images coupled with sparse depth maps obtained from a laser scanner mounted on a car. The images in this dataset have a resolution of $1216\times 352$. We utilize both the Eigen split \cite{eigen_depth} and the official split for our analysis. The Eigen split comprises 23158 training images and 697 test images, while the official split includes 42949 training images and 500 test images.

\paragraph{Metrics} Consistent with prior works \cite{newcrf, adabins, iebins}, we utilize a selection of well-established metrics to provide a comprehensive evaluation. The key metrics include: relative absolute error (Abs Rel), relative squared error (Sq Rel), root mean squared error (RMSE), absolute logarithmic error ($\text{log}_{10}$), root mean squared logarithmic error (RMSE log), inverse root mean squared error (iRMSE) and threshold accuracy ($\delta<1.25$, $\delta<1.25^2$, and $\delta<1.25^3$). Please refer to the appendix for details.

\begin{table*}
    \centering
    \footnotesize
    \renewcommand\arraystretch{1.25}
    \resizebox{0.95\linewidth}{!}{
        \begin{tabular}{c||c||cccc||ccc}
            \Xhline{1.5pt}
            Method        & Backbone      & Abs Rel $\downarrow$ & Sq Rel $\downarrow$ & RMSE $\downarrow$ & RMSE log $\downarrow$ & $\delta<1.25$ $\uparrow$ & $\delta<1.25^2$ $\uparrow$ & $\delta<1.25^3$ $\uparrow$ \\ \hline\hline
            BTS \cite{bts}                 & DenseNet-161  & 0.407          & 0.082          & 0.998          & 0.567          & 0.985          & {\ul 0.998}     & \textbf{1.000}  \\
            AdaBins \cite{adabins}         & E-B5+mini-ViT & 0.279          & 0.044          & 0.729          & 0.462          & 0.990          & {\ul 0.998}     & \textbf{1.000}  \\ \hline
            NeWCRFS \cite{newcrf}          & Swin-Large    & 0.533          & 0.244          & 1.004          & 0.792          & 0.956          & 0.976           & 0.988           \\
            PixelFormer \cite{pixelformer} & Swin-Large    & 0.534          & 0.230          & 1.076          & 0.782          & 0.957          & 0.979           & {\ul 0.991}     \\
            VA-DepthNet \cite{vadepth}     & Swin-Large    & {\ul 0.234}    & {\ul 0.029}    & {\ul 0.619}    & {\ul 0.373}    & \textbf{0.996} & \textbf{0.999}  & \textbf{1.000}  \\
            IEBins \cite{iebins}           & Swin-Large    & 0.528          & 0.238          & 0.999          & 0.790          & 0.956          & 0.976           & 0.988           \\ \hline
            \textbf{Ours}                  & Swin-Large    & \textbf{0.188} & \textbf{0.027} & \textbf{0.565} & \textbf{0.352} & {\ul 0.995}    & \textbf{0.999}  & \textbf{1.000}  \\
            \Xhline{1.5pt}
        \end{tabular}
    }
    \vskip -0.05in
    \caption{\textbf{Quantitative depth comparison on TOFDC dataset.} The maximum depth is capped at 5 meters. The first four error metrics are multiplied by 10 for presentation.}
    \label{tab.tofdc}
    \vskip -0.05in
\end{table*}

\begin{figure*}
    \centering
    \includegraphics[width=0.95\linewidth]{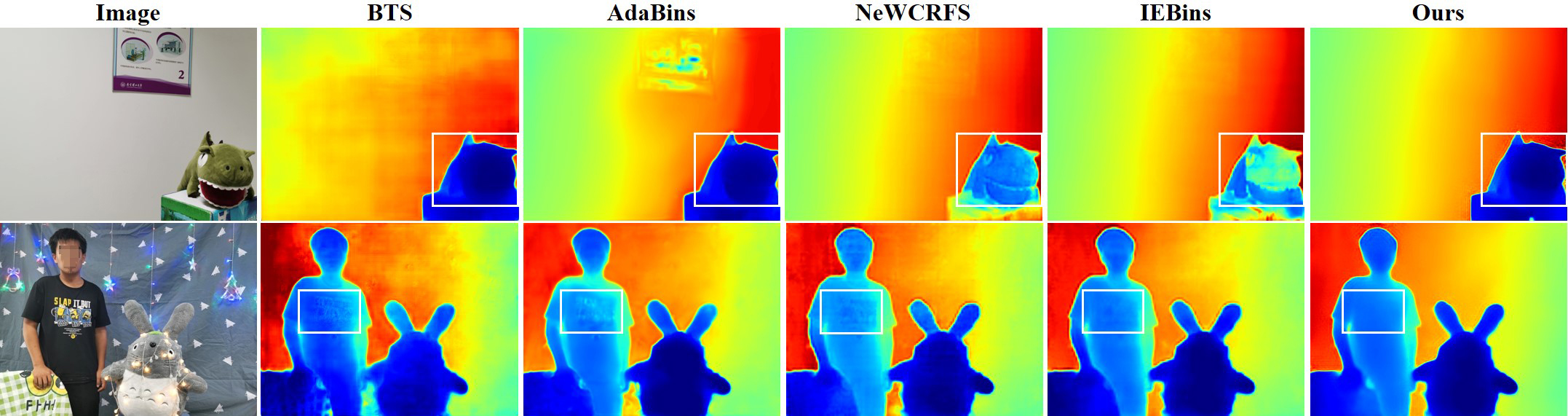}
    \vskip -0.05in
    \caption{\textbf{Qualitative depth comparison on the TOFDC dataset.}}
    \label{fig.tofdc}
    \vskip -0.05in
\end{figure*}

\subsection{Implementation Detail}

The DCDepth is implemented using Pytorch library \cite{pytorch}, and is trained with a batch size of 8 on four NVIDIA RTX-4090 GPUs with data-distributed parallel computing. Our method is trained on NYU-Depth-V2 dataset for 20 epochs, TOFDC dataset for 25 epochs, KITTI eigen split for 20 epochs and KITTI official split for 12 epochs. The optimization objective of our method is a combination of the scale-invariant log loss $L_d$, the frequency regularization $L_f$ and the smoothness regularization $L_s$, weighted by two scalar weights $\alpha$ and $\beta$:
\begin{equation}
    L = L_d + \alpha\cdot L_f+\beta\cdot L_s.
\end{equation}
For the NYU-Depth-V2 and TOFDC datasets, these two weights are set to $2\times 10^{-3}$ and 0.0, respectively, while for the KITTI dataset, both weights are set to $5\times 10^{-3}$. We opt for the Adam optimizer \cite{adam} and leverage the OneCycle learning rate scheduler \cite{onecycle}. The learning rate schedule entails an initial increase from $2\times 10^{-5}$ to $10^{-4}$ during the first 2 epochs, followed by a subsequent decrease to $5\times 10^{-6}$ using a cosine annealing strategy. To enhance generalization and mitigate overfitting, we integrate various data augmentation techniques into the training pipeline, including random horizontal flips, random rotations, random color jitter, and random image filtering. For feature extraction from images, we incorporate a Swin-Transformer architecture \cite{swin} pretrained on the ImageNet dataset \cite{imagenet} as the image encoder. To reduce the iteration steps necessitated for spectrum prediction, we further merge the frequency subgroups with indices $\{6,7\}$ and $\{8,\dots , 14\}$, leading to 9 iterative steps in total to generate the final depth predictions.

\subsection{Comparison with the State-of-the-Art}

\paragraph{NYU-Depth-V2} We benchmark our method against current \textbf{S}tate-\textbf{o}f-\textbf{T}he-\textbf{A}rt (SoTA) approaches on the indoor NYU-Depth-V2 dataset, with quantitative results presented in Tab. \ref{tab.nyu}. Despite vision transformers elevating the precision of depth estimation on this dataset, our method has surpassed existing SoTA approaches, particularly in the \emph{Abs Rel} and $\delta<1.25$ metrics. Qualitative comparisons, illustrated in Fig. \ref{fig.nyu}, reveal the adeptness of our method at capturing fine-grained geometries and producing smoother depth estimations in planar areas. Regions where our method outperforms are highlighted with white boxes, emphasizing its superior depth estimation accuracy.

\paragraph{TOFDC} The TOFDC dataset is characterized by its dense ground truth depth data. By utilizing this dataset, we demonstrate the enhanced capability of our method to effectively harness the dense ground truth, thereby achieving more accurate depth estimations compared to existing SoTAs. We present the quantitative results in Tab. \ref{tab.tofdc}, where our method demonstrates superior performance over existing SoTAs across a majority of the evaluated metrics. Specifically, our method achieves a significant improvement on the \emph{Abs Rel} and \emph{RMSE} metrics compared to VA-DepthNet, with enhancements of 19.7\% and 8.7\%, respectively. Fig. \ref{fig.tofdc} provides qualitative comparisons, illustrating that our method not only produces more accurate depth estimations but also more effectively delineates the object from the background, leading to more coherent depth estimations.

\begin{figure*}
    \centering
    \includegraphics[width=0.95\linewidth]{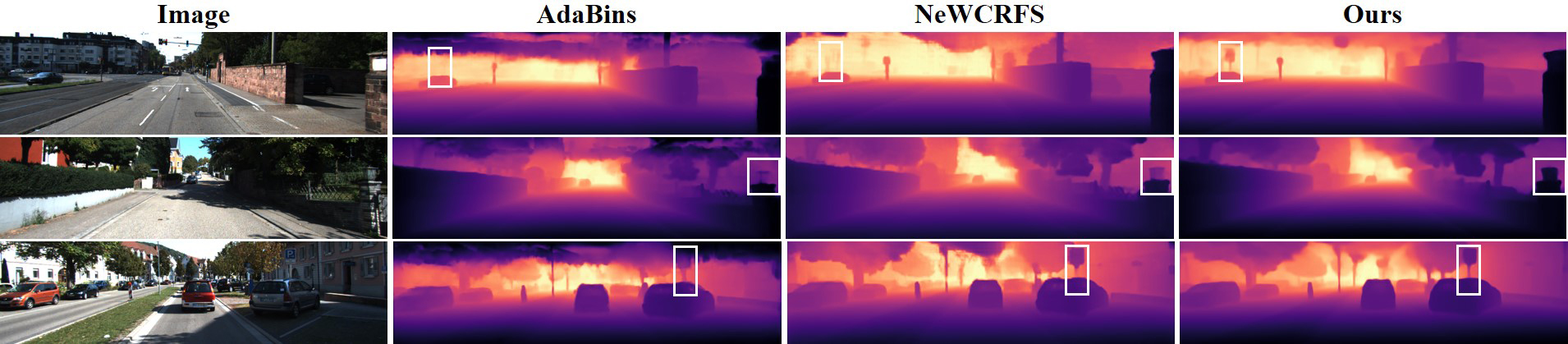}
    \vskip -0.05in
    \caption{\textbf{Qualitative depth comparison on the Eigen split of KITTI dataset.}}
    \label{fig.eigen}
    \vskip -0.05in
\end{figure*}

\paragraph{KITTI} We further evaluate our method on the outdoor dataset, KITTI, which has sparse depth ground truth collected with LiDAR. This sparsity presents a contrast to the denser depth information available in the NYU and TOFDC datasets, resulting in less robust supervision for learning frequency coefficients. Despite this challenge, our method demonstrates its robustness by achieving SoTA performance, which is attributed to the utilization of plenty training data coupled with our proposed regularization constraints. The quantitative analysis, as detailed in Tab. \ref{tab.eigen}, demonstrates the superior performance of our method. Qualitative evaluations, depicted in Fig. \ref{fig.eigen}, further substantiate the superiority of our method. The quantitative results on KITTI official split are reported in Tab. \ref{tab.kitti}. The pretrained weights from Semantic-SAM \cite{semantic_sam} are employed to initialize the encoder. Our method surpasses the compared approaches on the majority of metrics, particularly in the iRMSE metric, underscoring the robustness and effectiveness of our approach. 

\begin{table*}
    \centering
    \footnotesize
    \renewcommand\arraystretch{1.25}
    \resizebox{0.95\linewidth}{!}{
        \begin{tabular}{c||c||cccc||ccc}
            \Xhline{1.5pt}
            Method        & Backbone      & Abs Rel $\downarrow$ & Sq Rel $\downarrow$ & RMSE $\downarrow$ & RMSE log $\downarrow$ & $\delta<1.25$ $\uparrow$ & $\delta<1.25^2$ $\uparrow$ & $\delta<1.25^3$ $\uparrow$ \\ \hline\hline
            DORN \cite{fu2018deep}         & ResNet-101    & 0.072          & 0.307          & 2.727          & 0.120          & 0.932          & 0.984           & 0.994           \\
            VNL \cite{vnl}                 & ResNet-101    & 0.072          & –              & 3.258          & 0.117          & 0.938          & 0.990           & {\ul 0.998}     \\
            BTS \cite{bts}                 & DenseNet-161  & 0.060          & 0.249          & 2.798          & 0.096          & 0.955          & 0.993           & {\ul 0.998}     \\
            TransDepth \cite{transdepth}   & R-50+ViT-B/16 & 0.064          & 0.252          & 2.755          & 0.098          & 0.956          & 0.994           & \textbf{0.999}  \\
            AdaBins \cite{adabins}         & E-B5+mini-ViT & 0.058          & 0.190          & 2.360          & 0.088          & 0.964          & {\ul 0.995}     & \textbf{0.999}  \\
            P3Depth \cite{p3depth}         & ResNet-101    & 0.071          & 0.270          & 2.842          & 0.103          & 0.953          & 0.993           & {\ul 0.998}     \\ \hline
            NeWCRFS \cite{newcrf}          & Swin-Large    & 0.052          & 0.155          & 2.129          & 0.079          & 0.974          & \textbf{0.997}  & \textbf{0.999}  \\
            BinsFormer \cite{binsformer}   & Swin-Large    & 0.052          & 0.151          & 2.096          & 0.079          & 0.974          & \textbf{0.997}  & \textbf{0.999}  \\
            PixelFormer \cite{pixelformer} & Swin-Large    & {\ul 0.051}    & 0.149          & 2.081          & {\ul 0.077}    & {\ul 0.976}    & \textbf{0.997}  & \textbf{0.999}  \\
            VA-DepthNet \cite{vadepth}     & Swin-Large    & \textbf{0.050} & {\ul 0.148}    & 2.093          & \textbf{0.076} & \textbf{0.977} & \textbf{0.997}  & \textbf{0.999}  \\
            iDisc \cite{idisc}             & Swin-Large    & \textbf{0.050} & \textbf{0.145} & {\ul 2.067}    & {\ul 0.077}    & \textbf{0.977} & \textbf{0.997}  & \textbf{0.999}  \\ \hline
            \textbf{Ours}                  & Swin-Large    & {\ul 0.051}    & \textbf{0.145} & \textbf{2.044} & \textbf{0.076} & \textbf{0.977} & \textbf{0.997}  & \textbf{0.999} \\
            \Xhline{1.5pt}
        \end{tabular}
    }
    \vskip -0.05in
    \caption{\textbf{Quantitative depth comparison on the Eigen split of KITTI dataset.} The maximum depth value is capped at 80 meters.}
    \label{tab.eigen}
    \vskip -0.05in
\end{table*}

\paragraph{Parameter efficiency} We compare the parameter efficiency of our method with current SoTAs on the NYU-Depth-V2 dataset, with the input resolution set to $640\times 480$. The quantitative results, presented in Tab. \ref{tab.param}, reveal that our method exhibits the fewest training parameters while simultaneously achieving the best performance. For instance, our approach demonstrates a 9.0\% improvement in the \emph{RMSE} metric, while utilizing 4.1\% fewer parameters than NeWCRFS.

\subsection{Ablation Study}

We conduct comprehensive ablation studies to demonstrate the efficacy of the proposed PPH and PFF modules, and analyze the impact of the iteration steps on both model performance and inference speed. All experiments presented in this section are conducted on the NYU-Depth-V2 dataset.

\paragraph{Effect of PPH module} To assess the impact of the PPH module, we build a baseline by excluding the PPH from our method. In this setup, we employ a convolutional head to project the last-layer features to the output dimension. The final depth prediction is obtained through either bilinear and PixelShuffle \cite{pixelshuffle} upsampling or inverse DCT that converts the predicted frequency coefficients back to the spatial domain. Additionally, we introduce the adaptive bins \cite{adabins} as an alternative competitor. Quantitative experimental results are reported in Tab. \ref{tab.ablation_head}. Among the three approaches outputting in the spatial domain, the PixelShuffle-based approach performs the best. When predicting depth in the frequency domain, performance further improves, demonstrating the superiority of frequency-domain depth prediction. Lastly, our progressive prediction scheme significantly outperforms the compared approaches by a large margin, underscoring the efficacy of the PPH module.
%

\begin{table*}
    \centering
    \footnotesize
    \renewcommand\arraystretch{1.25}
    \resizebox{0.95\linewidth}{!}{
        \begin{tabular}{c||cc|ccccccc|c}
            \Xhline{1.5pt}
            Metric & \begin{tabular}[c]{@{}c@{}}DORN\\ \cite{fu2018deep}\end{tabular} & \begin{tabular}[c]{@{}c@{}}BTS\\ \cite{bts}\end{tabular} & \begin{tabular}[c]{@{}c@{}}NeWCRFS\\ \cite{newcrf}\end{tabular} & \begin{tabular}[c]{@{}c@{}}PixelFormer\\ \cite{pixelformer}\end{tabular} & \begin{tabular}[c]{@{}c@{}}BinsFormer\\ \cite{binsformer}\end{tabular} & \begin{tabular}[c]{@{}c@{}}iDisc\\ \cite{idisc}\end{tabular} & \begin{tabular}[c]{@{}c@{}}VA-DepthNet\\ \cite{vadepth}\end{tabular} & \begin{tabular}[c]{@{}c@{}}IEBins\\ \cite{iebins}\end{tabular} & \begin{tabular}[c]{@{}c@{}}NDDepth\\ \cite{nddepth}\end{tabular} & \textbf{Ours} \\ \hline\hline
            SILog $\downarrow$ & 11.77 & 11.67 & 10.39 & 10.28 & 10.14 & 9.89 & 9.84 & 9.63 & {\ul 9.62} & \textbf{9.60} \\
            Abs Rel $\downarrow$ & 8.78 & 9.04 & 8.37 & 8.16 & 8.23 & 8.11 & 7.96 & {\ul 7.82} & \textbf{7.75} & 7.83 \\
            Sq Rel $\downarrow$ & 2.23 & 2.21 & 1.83 & 1.82 & 1.69 & 1.77 & 1.66 & 1.60 & {\ul 1.59} & \textbf{1.54} \\
            iRMSE $\downarrow$ & 12.98 & 12.23 & 11.03 & 10.84 & 10.90 & 10.73 & 10.44 & 10.68 & {\ul 10.62} & \textbf{10.12} \\
            \Xhline{1.5pt}
        \end{tabular}
    }
    \vskip -0.05in
    \caption{\textbf{Quantitative depth comparison on the official split of KITTI dataset.} All metrics reported here are from the KITTI online leaderboard.}
    \label{tab.kitti}
    \vskip -0.05in
\end{table*}

\begin{table*}
    \centering
    \footnotesize
    \renewcommand\arraystretch{1.25}
    \resizebox{0.95\linewidth}{!}{
        \begin{tabular}{c||cccc|ccccc}
            \Xhline{1.5pt}
            &
            \multirow{2}{*}{\begin{tabular}[c]{@{}c@{}}NeWCRFS\\ \cite{newcrf}\end{tabular}} &
            \multirow{2}{*}{\begin{tabular}[c]{@{}c@{}}MG-Depth\\ \cite{mgdepth}\end{tabular}} &
            \multirow{2}{*}{\begin{tabular}[c]{@{}c@{}}IEBins\\ \cite{iebins}\end{tabular}} &
            \multirow{2}{*}{\begin{tabular}[c]{@{}c@{}}VA-DepthNet\\ \cite{vadepth}\end{tabular}} &
            \multicolumn{5}{c}{\textbf{Ours}} \\ \cline{6-10} 
            &                &       &       &                & 1 Step & 2 Steps & 3 Steps & 4 Steps & 9 Steps   \\ \hline\hline
            Param (M) $\downarrow$     & 270            & 296   & 273   & {\ul 262}      & \multicolumn{5}{c}{\textbf{259}}                                          \\ \cline{6-10} 
            Speed (FPS) $\uparrow$     & \textbf{37.95} & 24.24 & 21.51 & 15.68          & {\ul 31.55} & 28.72        & 26.03        & 24.07        & 14.24          \\ \hline
            RMSE $\downarrow$          & 0.334          & 0.311 & 0.314 & \textbf{0.304} & 0.310       & 0.307        & 0.306        & {\ul 0.305}  & \textbf{0.304} \\
            $\delta < 1.25$ $\uparrow$ & 0.922          & 0.933 & 0.936 & 0.937          & 0.937       & {\ul 0.939}  & {\ul 0.939}  & {\ul 0.939}  & \textbf{0.940} \\
            \Xhline{1.5pt}
        \end{tabular}
    }
    \vskip -0.05in
    \caption{\textbf{Parameter efficiency and inference speed on NYU-Depth-v2 dataset.} The right section enumerates the inference speed and corresponding performance metrics of our method at various iteration stages. All models are benchmarked on a single RTX 4090 GPU for consistency.}
    \label{tab.param}
    \vskip -0.05in
\end{table*}

\paragraph{Effect of PFF module} To evaluate the impact of the PFF module, we establish a baseline by excluding the PFF component from our method. We first introduce a convolutional layer and a PPM \cite{ppm} module to process the image feature at the last scale. Then, to validate the proposed DCT-based downsampling strategy, we replace it with bilinear and PixelUnshuffle \cite{pixelshuffle} downsampling. The quantitative experimental results are reported in Tab. \ref{tab.pff}. The first two approaches, which only process the last-scale feature, perform worse than the competitors with multi-scale feature aggregation. This demonstrates the necessity of multi-scale feature aggregation for depth prediction. Furthermore, our method, employing the DCT-based downsampling strategy, achieves the best performance, showcasing the effectiveness of our proposed DCT-based strategy for feature downsampling.

\paragraph{Effect of iterative steps} We analyze the impact of iterative steps on both prediction accuracy and inference speed. The results are reported in Tab. \ref{tab.param} and illustrated in Fig. \ref{fig.acc}. In summary, we observe that both prediction accuracy and inference time increase as the number of iterations grows. Leveraging the energy compaction property of the DCT, we strike a balance between accuracy and speed by selectively discarding predictions for high-frequency components. This strategic approach allows us to effectively reduce the number of iterative steps.

\begin{table*}
    \centering
    \footnotesize
    \renewcommand\arraystretch{1.25}
    \resizebox{0.95\linewidth}{!}{
        \begin{tabular}{c||c||ccc||ccc}
            \Xhline{1.5pt}
            
            Method        & Output Domain      & Abs Rel $\downarrow$ & Sq Rel $\downarrow$ & RMSE $\downarrow$ & $\delta<1.25$ $\uparrow$  & $\delta<1.25^2$ $\uparrow$ & $\delta<1.25^3$ $\uparrow$ \\ \hline\hline
            Baseline + Conv + Bilinear     & Spatial-Domain   & 0.090          & 0.042          & 0.319          & 0.929          & 0.991           & \textbf{0.998}  \\
            Baseline + AdaBins + Bilinear  & Spatial-Domain   & 0.088          & 0.042          & 0.319          & 0.932          & 0.991           & \textbf{0.998}       \\
            Baseline + Conv + PixelShuffle & Spatial-Domain   & 0.088          & 0.041          & 0.318          & 0.933          & \textbf{0.992}  & \textbf{0.998}  \\
            Baseline + Conv + inv DCT      & Frequency-Domain & 0.088          & 0.041          & 0.315          & 0.932          & \textbf{0.992}  & \textbf{0.998}  \\ \hline
            \textbf{Baseline + PPH}        & Frequency-Domain & \textbf{0.085} & \textbf{0.039} & \textbf{0.304} & \textbf{0.940} & \textbf{0.992}  & \textbf{0.998}  \\
            \Xhline{1.5pt}
        \end{tabular}
    }
    \vskip -0.05in
    \caption{\textbf{Ablation study on the PPH module.} The baseline is built by removing the PPH module. \emph{Conv} denotes linear projection with a convolutional layer. \emph{AdaBins} refers to the adaptive bins \cite{adabins}. All methods output at $\sfrac{1}{8}$ scale, and Bilinear and PixelShuffle \cite{pixelshuffle} are used to upsample the prediction.}
    \label{tab.ablation_head}
    \vskip -0.05in
\end{table*}

\begin{minipage}{\textwidth}
    \centering
    \begin{minipage}{0.61\textwidth}
        \centering
        \renewcommand\arraystretch{1.25}
        \resizebox{\linewidth}{!}{
            \begin{tabular}{c||cc|c}
                \Xhline{1.5pt}
                Method                          & Abs Rel $\downarrow$ & RMSE $\downarrow$ & $\delta<1.25$ $\uparrow$ \\ \hline\hline
                Baseline + Conv                 & 0.086          & 0.309          & 0.936          \\
                Baseline + PPM                  & 0.086          & 0.306          & 0.939          \\ \hline
                Baseline + PFF (Bilinear)       & \textbf{0.085} & 0.305          & \textbf{0.940} \\
                Baseline + PFF (PixelUnshuffle) & \textbf{0.085} & 0.306          & \textbf{0.940} \\ \hline
                \textbf{Ours}                   & \textbf{0.085} & \textbf{0.304} & \textbf{0.940} \\
                \Xhline{1.5pt}
            \end{tabular}
        }
        \makeatletter\def\@captype{table}\makeatother\caption{\textbf{Ablation study on the PFF module.} The baseline is built by removing the PFF module. We evaluate the proposed DCT-based downsampling strategy by replacing it with bilinear and PixelUnshuffle \cite{pixelshuffle} downsampling.}
        \label{tab.pff}
    \end{minipage}
    \hspace{0.1in}
    \begin{minipage}{0.36\textwidth}
        \centering
        \includegraphics[width=\linewidth]{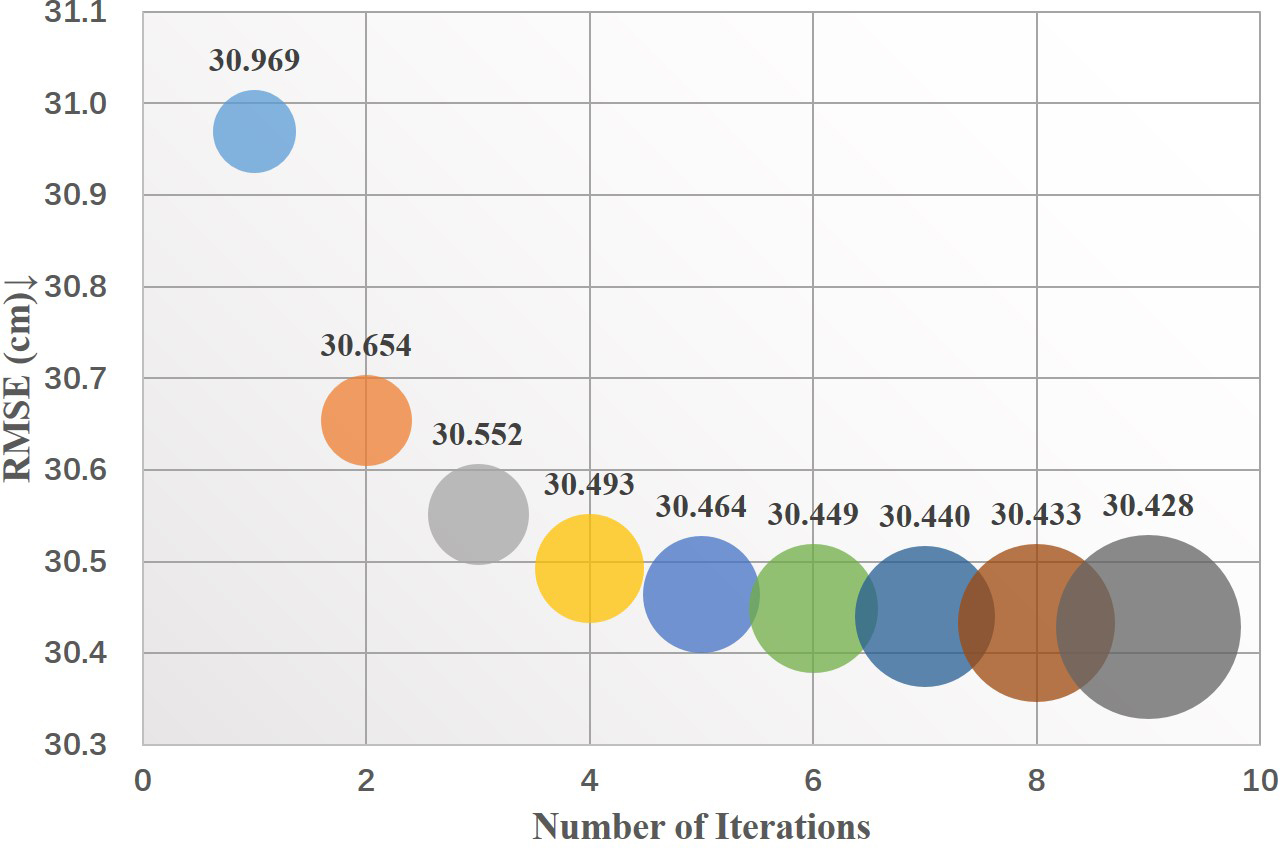}
        \vskip -0.1in
        \makeatletter\def\@captype{figure}\makeatother\caption{\textbf{Accuracy vs. inference speed.} The width of each bubble corresponds to the processing time.}
        \label{fig.acc}
    \end{minipage}
\end{minipage}

\section{Limitation and Broader Impact}

Our method employs the differentiable inverse DCT to transform the predicted spectrum back to the spatial domain. By minimizing the difference between the spatial-domain estimation and the valid ground truth, our model can be trained end-to-end. However, the sparsity of the ground truth may lead to inefficient supervision of the frequency estimation. While we have proposed two regularization terms to prevent our model from being incorrectly optimized, we observe that our method is more effective with dense supervision. Exploring more effective training strategies when only sparse depth ground truth is available will be an important research direction for our future work.

Monocular depth estimation is a pivotal technique for interpreting 3D scenes from 2D images and has widespread applications in autonomous driving, robotics, and 3D modeling, among others. Given the extensive applications of this task, our method is poised to positively impact these fields by advancing their capabilities. Considering the fundamental nature of monocular depth estimation, our work is not anticipated to have a significant negative societal impact.

\section{Conclusion}

In this paper, we introduce DCDepth, a novel framework for the MDE task. Departing from existing methods, our method progressively estimates patch-wise depth in the frequency domain and then recovers spatial-domain depth via inverse DCT. This formulation inherently models local depth correlations and frames the estimation process as a global-to-local scheme, achieving more accurate depth estimation. Leveraging the energy compaction property of DCT, our method strikes an effective balance between accuracy and inference speed, making it well-suited for practical applications.

\section{Acknowledgment}

We would like to thank the reviewers and the chairs for their suggestions and efforts. This work was partially supported by the National Natural Science Foundation of China under Grant 62361166670 and 62072242, the Fundamental Research Funds for the Central Universities under Grant 070-63233084, the Young Scientists Fund of the National Natural Science Foundation of China under Grant 62206134 and the Tianjin Key Laboratory of Visual Computing and Intelligent Perception. The PCA Lab is associated with the Key Lab of Intelligent Perception and Systems for High-Dimensional Information of Ministry of Education, and Jiangsu Key Lab of Image and Video Understanding for Social Security, School of Computer Science and Engineering, Nanjing University of Sci \& Tech.

{   
    \bibliographystyle{ieeenat_fullname}
    \bibliography{references}
}


\end{document}